\documentclass[sn-basic]{sn-jnl} 

\usepackage{graphicx}%
\usepackage{multirow}%
\usepackage{amsmath,amssymb,amsfonts}%
\usepackage{amsthm}%
\usepackage{mathrsfs}%
\usepackage[title]{appendix}%
\usepackage{xcolor}%
\usepackage{textcomp}%
\usepackage{manyfoot}%
\usepackage{booktabs}%
\usepackage{algorithm}%
\usepackage{algorithmicx}%
\usepackage{algpseudocode}%
\usepackage{listings}%
\usepackage{tcolorbox}

\usepackage{subfig}
\usepackage{graphicx}
\usepackage{subcaption}

\theoremstyle{thmstyleone}%
\theoremstyle{thmstyletwo}%

\theoremstyle{thmstylethree}%

\begin{document}

\title[Article Title]{Advancing Minority Stress Detection with Transformers: Insights from the Social Media Datasets}


\author[1]{\fnm{Santosh} \sur{Chapagain}}\email{santosh.chapagain@usu.edu}

\author[2]{\fnm{Cory J} \sur{Cascalheira}}\email{cjcascalheira@gmail.com}

\author[1]{\fnm{Shah Muhammad} \sur{Hamdi}}\email{s.hamdi@usu.edu}

\author[1]{\fnm{Soukaina Filali} \sur{Boubrahimi}}\email{soukaina.boubrahimi@usu.edu}

\author[3]{\fnm{Jillian R.} \sur{Scheer}}\email{jrscheer@syr.edu}

\affil*[1]{\orgdiv{Department of Computer Science}, \orgname{Utah State University}, \orgaddress{\street{4205 Old Main Hill}, \city{Logan}, \postcode{84322}, \state{UT}, \country{USA}}}

\affil[2]{\orgdiv{VA Puget Sound Health Care System}, \orgname{Seattle VA Medical Center}, \orgaddress{\street{1660 South Columbian Way}, \city{Seattle}, \postcode{98108}, \state{WA}, \country{USA}}}

\affil[3]{\orgdiv{Department of Psychology}, \orgname{University of Rhode Island}, \orgaddress{\street{142 Flagg Road}, \city{Kingston}, \postcode{02881}, \state{RI}, \country{USA}}}


\abstract{Individuals from sexual and gender minority groups experience disproportionately high rates of poor health outcomes and mental disorders compared to their heterosexual and cisgender counterparts, largely as a consequence of minority stress as described by Meyer’s (2003) model. This study presents the first comprehensive evaluation of transformer-based architectures for detecting minority stress in online discourse. We benchmark multiple transformer models including ELECTRA, BERT, RoBERTa, and BART against traditional machine learning baselines and graph-augmented variants. We further assess zero-shot and few-shot learning paradigms to assess their applicability on underrepresented datasets. Experiments are conducted on the two largest publicly available Reddit corpora for minority stress detection, comprising 12,645 and 5,789 posts, and are repeated over five random seeds to ensure robustness. Our results demonstrate that integrating graph structure consistently improves detection performance across transformer-only models and that supervised fine-tuning with relational context outperforms zero and few-shot approaches. Theoretical analysis reveals that modeling social connectivity and conversational context via graph augmentation sharpens the models’ ability to identify key linguistic markers such as identity concealment, internalized stigma, and calls for support, suggesting that graph-enhanced transformers offer the most reliable foundation for digital health interventions and public health policy.}




\maketitle

\section{Introduction}\label{sec1}
Stress is a state of worry or mental tension caused by a difficult situation. \textit{Minority stress} is a specific type of stress that disproportionately affects marginalized groups \textemdash communities discriminated against based on race, sex, sexual orientation or disability \citep{meyer2003prejudice}. \textit{Minority stress} is defined as the high levels of stress experienced by stigmatized groups due to factors such as poor social support, low socioeconomic status, prejudice, and discrimination \citep{meyer2003prejudice}. Lesbian, gay, bisexual, transgender, queer, and other people with sexual and gender minorities (LGBTQ+) are especially vulnerable among marginalized groups \citep{salerno2020lgbtq}. \textit{Minority stress} plays a major role in health disparities \citep{pachankis2020sexual, de2022minority} between them and the majority population. Research shows that these forms of stress cause serious mental and physical health harm that includes anxiety, depression, and, in extreme cases, suicidal ideation \citep{de2022minority}. \textit{Minority stress} theory \citep{meyer2003prejudice} explains how adverse social conditions contribute to chronic stress and health disparities. Some of the key contributors \citep{cascalheira2023predicting} to minority stress include (1) prejudiced events \textemdash discrimination and violence, (2) expected rejection \textemdash anticipate discrimination even in the absence of it, (3) identity concealment \textemdash concealing one's sexuality to avoid discrimination, (4) internalized stigma \textemdash feeling shame and a desire to hide their identity, (5) gender dysphoria \textemdash distress from a mismatch between gender identity and birth sex.

Although the impact of minority stress is well known, its detection remains challenging due to its linguistic complexity \citep{cascalheira2023predicting}. In their paper, \cite{cascalheira2023predicting} explain how minority stress is characterized by linguistic sophistication, involving advanced syntactic, semantic, and pragmatic elements. It serves as a social determinant of health disparities due to its unique linguistic features:

\begin{itemize}
\item \textbf{Specific semantics and pragmatics}: Minority stress expressions often incorporate cultural idioms unique to LGBTQ+ communities, such as “I’m not out” or “I’m in the closet,” to convey identity concealment.

\item \textbf{Psycholinguistic permutations}: The same stress factor, like internalized stigma, can be expressed in varied ways, e.g., “I don’t like who I am” or “bisexual people just can’t make up their mind.”

\item \textbf{Lexical density}: Communicating minority stress requires a detailed context to distinguish it from general stress or other health issues, highlighting the complexity of accurately interpreting such expressions.
\end{itemize}

Its subtle and complex linguistic expressions, often marked by sarcasm, ambiguity, or implicit references, are difficult to identify using traditional methods. Addressing this issue requires the development of robust datasets and advanced models capable of detecting minority stress in textual data. Such tools could enable early interventions, provide crucial support for at-risk individuals, and inform strategies to reduce the harmful effects of minority stress. 

Social media platforms, where LGBTQ+ individuals often disclose their experiences, have become valuable sources for studying linguistic expressions of minority stress \citep{durstewitz2019deep, selkie2020transgender,woznicki2021parasocial}. Reddit and its subreddits are ideal for studying minority stress because they provide access to diverse, authentic, and anonymous discussions in targeted communities (e.g. \textit{r/lgbt}). These platforms capture culturally specific language, emotional expressions, and personal narratives, offering rich, real-time, and publicly accessible data to analyze how minority stress manifests across various identities and contexts. As a result, Reddit has become an essential space for LGBTQ+ individuals to connect, come out, and seek support while minimizing the risk of prejudice \citep{mcdermott2015asking, mcdermott2012youth}.

Previous research has explored the use of natural language processing (NLP) techniques, including hybrid models, for the classification of minority stress \citep{cascalheira2022classifying,saha2019language, cascalheira2023predicting, cascalheira2024lgbtq+, chapagain2024predictive}. Still, it lacks a fair comparison between transformer-based models and only uses one dataset. Building on this foundation, our study offers the first comprehensive evaluation of transformer-based models for minority stress classification. Specifically, this paper makes the following contributions:
\begin{itemize}
    \item We introduce and empirically test the hypothesis that graph-augmented transformers (e.g., BERT-GCN, RoBERTa-GCN, BART-GCN) outperform sequence-only models by capturing community co-occurrence patterns, conversational context, and distinctive linguistic features of minority stress.
    \item We evaluate a range of transformer-only architectures, including BART, ELECTRA, BERT, RoBERTa, and GPT-2, on the two largest publicly available Reddit corpora for minority stress detection (12,645 and 5,789 posts), reporting robust results averaged over five random seeds.
    \item We assess these graph-augmented variants on how relational edges (user and thread connections) enhance detection by modeling homophily, discourse coherence, cultural idioms (e.g., “I’m in the closet”), and psycholinguistic permutations (e.g., “I don’t like who I am” vs. “bisexual people just can’t make up their mind”).
    \item We explore zero-shot and few-shot classification with ChatGPT to evaluate the generalizability of large pre-trained models on underrepresented datasets, and we compare these results against supervised fine-tuning.
    \item We show that while zero and few shot methods show promise in low-resource scenarios, supervised fine-tuning, especially with graph augmentation, consistently achieves superior performance, highlighting the importance of labeled data and relational context for reliable minority stress detection.
\end{itemize}

\section{Related Work}
Recent research has increasingly focused on analyzing textual data\textemdash such as social media posts, interviews, and survey responses\textemdash to classify and interpret minority stress expressions. The theoretical foundational work of authors like \cite{meyer2003prejudice}, who introduced the minority stress model, has been extended by empirical studies applying computational methods. For example, \cite{pachankis2007psychological} and \cite{hendricks2012conceptual} explored the psychological mechanisms of minority stress and its impact on mental health outcomes, providing a foundation for further exploration using NLP. 

Machine learning classifiers have been widely used to detect expressions of depression, suicidal ideation, and other mental health concerns in online communities \citep{bagroy2017social,de2014mental, de2016discovering, saha2019language}. For example, researchers have found that self-disclosure posts in communities like \textit{r/SuicideWatch} and \textit{r/depression} align with clinical descriptions of mental health problems, providing valuable information on stigmatized experiences \citep{de2014mental, saha2019language}. With advances in NLP, many researchers \citep{yang2023automatic} have analyzed and classified stress-related expressions in text using techniques such as sentiment analysis, topic modeling, and machine learning. Similarly, deep learning models have proven effective in predicting social determinants of health disparities, including cyberbullying and racism \citep{shaw2022investigations}, as well as hate speech detection on larger datasets \citep{chapagain2025advancinghatespeechdetection, piot2024metahate}. Hybrid models, such as BERT-LSTM, have also shown strong performance in predicting misogynistic speech on Twitter \citep{angeline2022misogyny}. Some recent work has also examined bias and harmful speech targeting LGBTQ+ communities, including gender neutral form integration for bias reduction~\citep{sobhani2023measuring}, cyberbullying detection specific to LGBTQ+ contexts~\citep{arslan2024detecting}, and harmful conversational content classification~\citep{dacon2022detecting}.

Online discussions have become a valuable resource for examining minority stress, addressing the limitations of traditional research methods based on convenience-based samples, such as surveys \citep{martinez2014still, nguyen2016negative}. Social media platforms, especially Reddit, are useful for reaching minority populations because their anonymity encourages self-disclosure and the creation of communities around stigmatized topics, such as \textit{r/lgbtq+}, \textit{r/asktransgender}, and \textit{r/queer} \citep{andalibi2018social, andalibi2016understanding, saha2019language}.

\cite{saha2019language} utilized Reddit as a social media platform to study self-disclosure expressions of minority stress among LGBTQ+ individuals, identifying it as a scalable and inclusive method to capture diverse experiences in geographies, sexual orientations, and gender identities. They used traditional machine learning models, including logistic regression and multilayer perceptron, achieving a best F1 score of 0.75, although the model struggled with sequential data. \cite{cascalheira2022classifying} introduced a Bi-LSTM model but, due to architectural limitations, achieved only an F1 score of 0.61. Subsequently, the combination of BERT-CNN demonstrated a strong performance in predicting both composite minority stress and its factors \citep{cascalheira2023predicting}. \cite{chapagain2024predictive} utilized transductive learning techniques to analyze the disclosure of social networks, uncovering predictive patterns of minority stress expressions and their correlation with mental health outcomes among LGBTQ+ individuals. However, the transductive nature of their model limits its ability to generalize and handle unseen data, restricting its applicability to new or evolving contexts beyond the specific dataset used for analysis.

In contrast to previous research, our research addresses the limitations of previous work by providing a comprehensive evaluation of transformer-based models for minority stress classification. We compare both sequence-only models (ELECTRA, BERT, RoBERTa, BART) and their graph-augmented variants across two only two publicly available Reddit datasets (12,645 and 5,789 posts). We also investigate zero-shot and few-shot classification to assess how well pre-trained models generalize to underrepresented data with limited labels. Our results demonstrate that incorporating graph structure consistently enhances detection performance, confirming that graph-augmented transformers more effectively identify nuanced markers of minority stress than sequence-only approaches.

By bridging computational linguistics with psychological theories, this growing body of research continues to illuminate the multifaceted experiences of minority stress among LGBTQ+ individuals, offering transformative potential for academic understanding and practical interventions.

\section{Methodology}\label{sec3}
 
\subsection{Transformer and Graph Convolutional Network}  
The BERT-GCN framework Figure 1 integrates BERT to initialize the embeddings of document nodes in a heterogeneous Reddit corpus graph. These embeddings are then processed through a graph convolutional network (GCN), iteratively refining node representations. The final outputs are passed through a linear transformation and a softmax layer for classification. 

\begin{figure*}[!htbp]
\centerline{\includegraphics[width=0.8\textwidth]{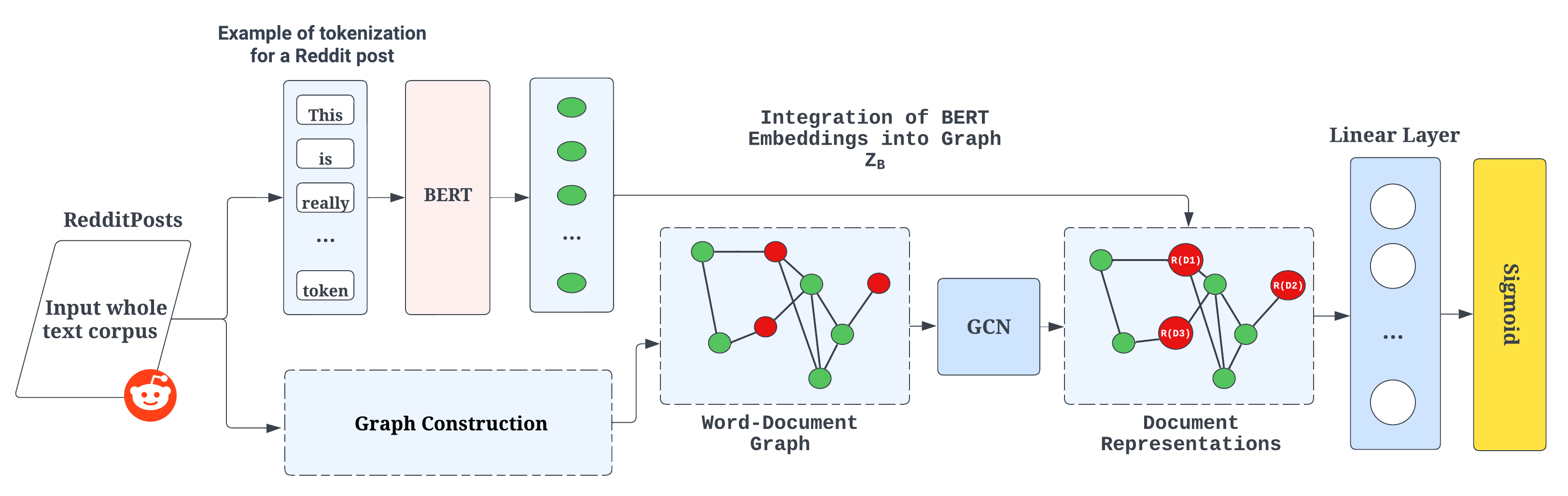}}
\caption{BERT-GCN Network Architecture.}
\label{architecture1}
\end{figure*}

In our graph representation, the document nodes ($N_{doc}$) and word nodes ($N_{word}$) form the initial feature matrix and is given by below equation (1):  

\begin{equation}  
X = \begin{pmatrix}  
Z_{doc} \\  
0  
\end{pmatrix}_{(N_{doc}+N_{word}) \times d}  
\end{equation}  

where $Z_{doc}$ denotes document embeddings. These embeddings undergo transformation via:  

\begin{equation}  
Z_B = \text{softmax}(W X)  
\end{equation}  

where $W$ is the weight matrix.  

To construct a heterogeneous graph, we follow TextGCN \citep{yao2019graph} and BertGCN \citep{lin2021bertgcn}, using term frequency-inverse document frequency (TF-IDF) for document-word edges and positive pointwise mutual information (PPMI) for word-word relationships. The edge weights between nodes $i$ and $j$ are assigned as follows:  

\begin{equation}  
A_{i,j} =  
\begin{cases}  
\text{TF-IDF}(i, j), & \text{if } i \text{ is a document, } j \text{ is a word} \\  
\text{PPMI}(i, j), & \text{if } i, j \text{ are words and } i \neq j \\  
1, & \text{if } i = j \\  
0, & \text{otherwise}  
\end{cases}  
\end{equation}  

PPMI measures the correlation between words:  

\begin{equation}  
\text{PPMI}(i, j) = \log \frac{p(i, j)}{p(i)p(j)}  
\end{equation}  

Here, \( p(i, j) \) and \( p(i) \) represent the probabilities of the word pair \( (i, j) \) and the single word \( i \) occurring in the corpus, respectively. The terms \( \#W(i, j) \) and \( \#W(i) \) refer to the number of sliding windows that contain both words \( (i, j) \) and the word \( i \), respectively. Finally, \( \#W \) denotes the total number of sliding windows in the corpus.

Edges are added only for positive PPMI values, ensuring meaningful semantic relationships. The input matrix $X$ is then passed through the GCN layers, with the output defined as:  

\begin{equation}  
Z_G = \text{softmax}(\text{gcn}(X, \widetilde{A}))  
\end{equation}

The BERT-GCN model combines predictions from both BERT and GCN using linear interpolation.  

\begin{equation}  
Z_{Final} = \lambda Z_G + (1 - \lambda) Z_B  
\end{equation}  

where $\lambda \in (0,1)$ controls the balance between the two models. When $\lambda = 0$, the model is entirely based on BERT, while $\lambda = 1$ prioritizes GCN. This adaptive weighting optimizes classification performance based on the characteristics of the dataset.

\subsection{Transformer and Convolutional Neural Network} 
The BERT-CNN model, shown in Figure 2 consists of four main layers: the BERT embedding layer, the convolutional layer, the fully connected neural network (FCNN) and the sigmoid layer. The model processes text sequences by converting them into 768-dimensional contextual word embeddings using BERT, which are then passed through three convolutional layers, each with 100 filters and kernel sizes of [3, 4, 5] to capture diverse text patterns. Max pooling extracts key features that are concatenated and pass through a dropout layer (0.5) before reaching the FCNN and the sigmoid layer for classification. The model is trained for 10 epochs with a batch size of 32, using binary cross-entropy with logits as the loss function. The Adam optimizer, with a learning rate of \( \alpha = 10^{-3} \), momentum hyperparameter \( \beta_1 = 0.9 \), RMS-prop hyperparameter \( \beta_2 = 0.999 \), and the weight decay hyperparameter, updates model parameters efficiently. The sigmoid layer converts raw scores into probabilities, applying a threshold of 0.5 for final binary classification.

\begin{figure*}[htbp]
\centerline{\includegraphics[width=0.8\textwidth]{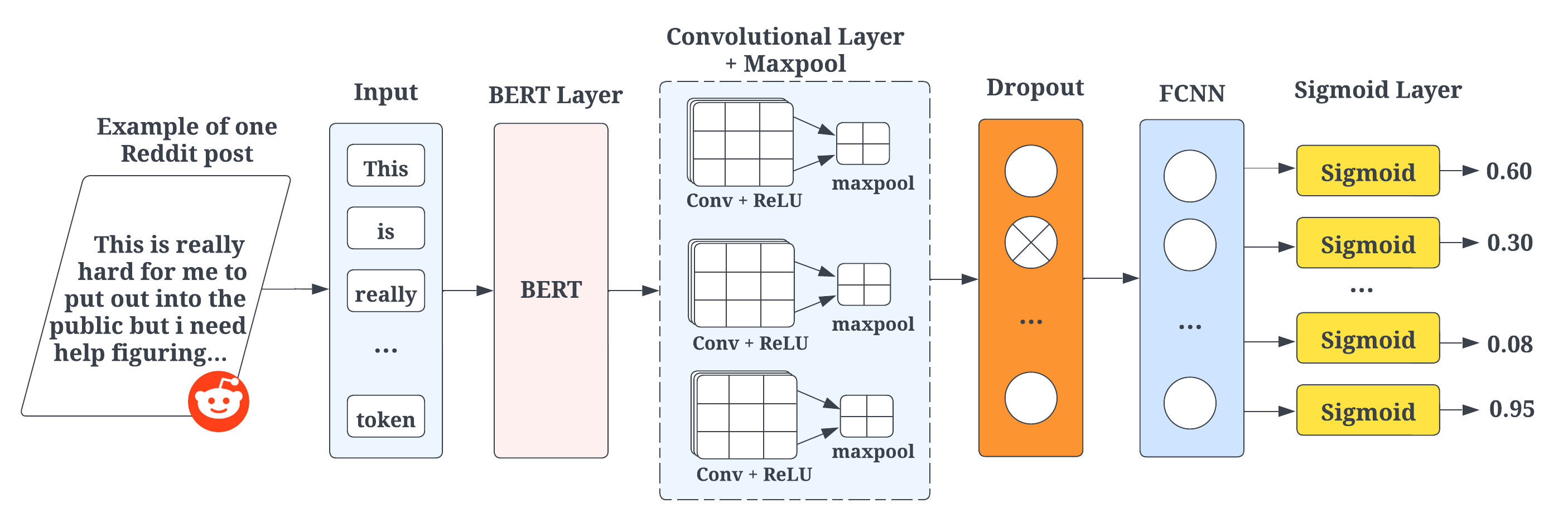}}
\caption{BERT-CNN network architecture}
\label{architecture1}
\end{figure*}

\subsection{Transformer and Gated Recurrent Unit
Model}\label{subsec2}

As shown in Figure 2, this architecture modifies the BERT-CNN model by replacing the convolutional layer with a BiGRU layer. The step-by-step learning process for the BERT-BiGRU framework, which consists of four distinct layers: the BERT layer, the BiGRU layer, the fully connected neural network layer (FCNN), and a sigmoid layer. The process begins by taking a batch of text sequences as input and converting them into contextual word embeddings using BERT. These embeddings serve as the initial input for the BiGRU layer, which captures sequential information both in the forward and backward directions. Subsequently, dropout regularization is applied to the output of BiGRU, which is then passed through the linear layer and a sigmoid layer for binary classification.

\subsection{Transformer-Based Models with Linear Layers}

\begin{figure}[!htbp]
\centerline{\includegraphics[width=8cm]{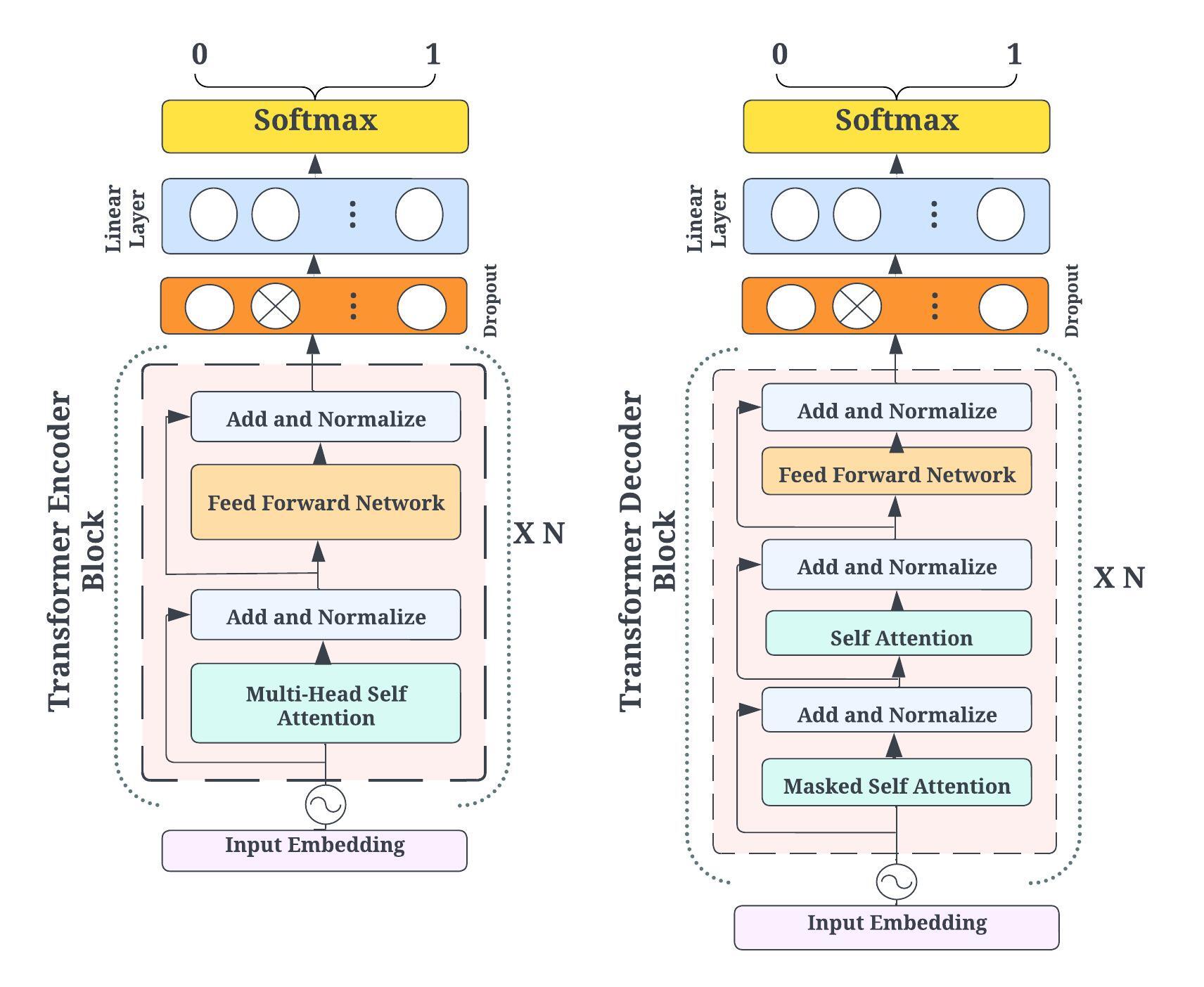}}
\caption{Encoder-Based (Left) and Decoder-Based (Right) Transformer Architectures}
\label{transformer-model}
\end{figure}

This section describes how transformer-based architectures—particularly those leveraging attention mechanisms—are applied to minority stress detection. First introduced by \cite{vaswani2017attention}, transformers replace recurrent or convolutional structures with multi-head self-attention, allowing for parallel processing and richer contextual embeddings. Figure 3 illustrates the two main variants of the transformer architecture:

\begin{itemize}
\item Encoder-based transformers (e.g. BERT) use a bidirectional design. Tokens are embedded with positional information, and each layer uses multi-head self-attention plus feed-forward networks. The residual and normalization layers stabilize the training, and the final output is fed through the classification layers.

\item Decoder-based transformers (e.g., GPT) use a unidirectional (causal) mask in self-attention, focusing only on past tokens. After each self-attention step, a feed-forward network refines the representations and the final outputs go through softmax.

For our minority stress classification task on both datasets, we used the base model for all transformers to ensure a fair comparison. We set the maximum sequence length to 512 and trained the model for 4 epochs using a batch size of 4 with a learning rate of 3e-5.

\end{itemize}

\subsubsection{BERT}  
BERT (Bidirectional Encoder Representations from Transformers) \citep{devlin2018bert} is a transformer-based model designed to understand context by processing text bidirectionally. It was trained on large datasets like the Toronto Book Corpus and Wikipedia using masked language modeling and next-sentence prediction. These techniques allow BERT to grasp meaning more effectively, making it a strong performer in various natural language processing (NLP) tasks.

\subsubsection{RoBERTa}  
RoBERTa (Robustly Optimized BERT Pretraining Approach) \citep{liu2019roberta} enhances BERT by removing the next-sentence prediction task and focusing on more effective masked language modeling. With larger batch sizes, longer training times, and dynamic masking, RoBERTa improves both performance and robustness, making it a strong alternative to BERT.

\subsubsection{ELECTRA}  
ELECTRA (Efficiently Learning an Encoder that Classifies Token Replacements Accurately) \citep{clark2020electra} introduces a more efficient pre-training approach. Instead of simply masking words like BERT, it trains two networks: a generator, which replaces words, and a discriminator, which learns to detect these replacements. This method significantly improves token-level understanding while being computationally efficient.

\subsubsection{GPT-2}  
GPT-2 (Generative Pre-trained Transformer 2) \citep{radford2019language} is an autoregressive transformer model with 1.5 billion parameters, trained on a diverse 40GB dataset. Unlike BERT, GPT-2 is unidirectional, meaning that it generates text one token at a time. Using Byte-Pair Encoding for efficient text processing, GPT-2 is particularly powerful for text generation tasks such as dialogue and storytelling.

\subsubsection{BART}  
BART (Bidirectional and Auto-Regressive Transformer) \citep{lewis2019bart} blends the best of both worlds by combining a bidirectional encoder (like BERT) with an autoregressive decoder (like GPT). This hybrid architecture allows BART to capture complex contextual relationships while generating fluent and coherent text, making it highly effective for tasks such as summarization and translation.

\subsubsection{DeBERTa}  
DeBERTa (Decoding-enhanced BERT with Disentangled Attention) \citep{he2020deberta} builds on BERT and RoBERTa by introducing a disentangled attention mechanism, which separates content from positional information. In addition, an improved mask decoder helps it better capture contextual relationships, resulting in enhanced language understanding.

\subsubsection{Longformer}  
Longformer \citep{beltagy2020longformer} is designed to efficiently handle long documents, overcoming the memory limitations of traditional transformers. This is achieved through a sparse attention mechanism that combines local and global attention patterns. This allows Longformer to process extended sequences effectively, making it particularly useful for document summarization and other long-form text applications.

\subsection{Zero-Shot Classification}
Zero-shot learning is an NLP technique that classifies data into categories without explicit training, enabling models to predict unseen classes \citep{xian2018zero}. In this work, we use a Zero-shot Learning approach for minority stress classification. The model used a pre-trained language model, which can be thought of as an example of transfer learning \textemdash applying the knowledge of a pre-trained model to a new related task. In this setting, we provide the prompt and the sequence of text as an instruction for what we want our model to do. Zero-shot classification differs from single-shot or few-shot classification as it does not use any examples of the target task, unlike the latter, which uses one or a few examples. These capabilities emerge in large-language models with more than 100M parameters, and their effectiveness improves as the model size increases, with larger models generally performing better at these tasks. An example of a zero-shot prompt for classifying whether there is a presence of minority stress or not from a sequence of text is shown below:
\begin{tcolorbox}[colback=blue!5!white,colframe=blue!75!black]
\textbf{Task:} Classify the following input text into one of the following two categories: [\textit{minority stress, no minority stress}] \\

\textbf{Input Text:} \\
\textit{I have to be straight if I want things in life. Being a lesbian will mean having a life where everything I want will be extremely hard to get.} \\

\textbf{Output:} \\
\textit{minority stress}
\end{tcolorbox}

\subsection{Few-Shot Classification}
Few-shot learning is a newer approach that enables model training with fewer labeled data. For our implementation, we used 3-shot and 10-shot learning using ChatGPT (GPT-4o) for the classification of minority stress text. For 3-shot learning, we provided the model with two positive-labeled examples and one negative-labeled example before performing classification on a test set consisting of 869 samples from the MiSSOM+ dataset. Similarly, in the 10-shot learning approach, we increased the number of training examples by providing five positive and five negative labeled samples.  By implementing 3- and 10-shot learning, we aimed to assess how the number of provided examples influences model performance and whether increasing the number of labeled instances leads to a significant improvement in classification accuracy.

\section{Experiments}
For implementing the model, we used Python 3.10.12, Numpy 1.22.4, Pandas 1.5.3, and PyTorch 1.13.0 with Cuda 11.1. We ran all the experiments on the Linux server, which supports dual Intel Xeon Gold 5220R processors, each with 24 cores at 2.20 GHz, with a substantial 35.75 MB cache. The source code can be found on GitHub.\footnote{\url{https://github.com/chapagaisa/transformer}}. The Institutional Review Board of the university of the second author approved this study as exempt. Scores presented on results (Section 5) are averages of 5 runs with different seeds and subscriptions indicate standard deviation. 

\subsection{Dataset}\label{subsec2}
In this study, we used two Reddit-based datasets. Both datasets have access control requirements, although they are free to use after approval from the authors. The datasets scraped data from Reddit, and informed consent was neither required nor obtained. Personal identifiers (i.e., Reddit usernames) were not included in the datasets. We maintained the datasets on password-protected servers during the experiments. For more information on access control and consent, see Ethical Considerations. The two datasets were split into 70\% for the training set, 15\% for the validation set, and 15\% for the test set. In each dataset, minority stress was coded as a binary outcome, where 1 represents the presence of minority stress and 0 represents the absence of minority stress. Table 1 shows the text examples of each label, with the text slightly paraphrased to protect the anonymity of the Reddit user.

The first dataset was made available by \cite{saha2019language} and contained 12,645 examples of minority stress on Reddit, of which 350 were human-annotated and 12,295 were machine-annotated. All Reddit posts were scraped from subreddit r/lgbt. Saha et al. developed a codebook by annotating 50 posts to achieve consensus between two coders. A single coder annotated 300 additional posts independently. The team trained MLP on these 350 posts and then used MLP to machine-annotate the remaining posts.

Because machine annotation can introduce labeling errors, we used a second, newer dataset: the LGBTQ+ Minority Stress on Social Media (MiSSoM+) dataset \citep{cascalheira2024lgbtq+}, which contains 5,789 human-annotated Reddit posts related to minority stress. Reddit posts were scraped from the following subreddits: r/actuallesbians, r/ainbow, r/bisexual, r/gay, r/genderqueer, r/questioning, and r/trans. The MiSSoM+ team, consisting of mental health clinicians, psychologists, and students with expertise in LGBTQ+ health, used conventional content analysis to annotate the posts. They developed a codebook and annotated 375 posts to achieve consensus among nine coders. The nine coders then independently coded the remaining posts. Three additional coders periodically assisted with annotation under the supervision of the lead author \citep{cascalheira2024lgbtq+}. The team met weekly to discuss their coding decisions, engage in peer debriefing, and audit one another’s annotation decisions, resolving disagreements through consensus.

The MiSSoM+ dataset comprises 4,551 positive class labels and 1,238 negative class labels, while the Saha dataset contains 8,264 positive and 4,323 negative class labels. Here, the label 1 represents the presence of minority stress, and 0 indicates its absence. Both datasets are imbalanced, which might lead the model to be biased toward the majority class. To overcome this imbalance problem, we used stratification. By using stratified sampling, the imbalanced classes are represented proportionally in each subset of the data, enabling a more reliable evaluation of the model's performance across different classes.

\begin{table}[htbp]
  \centering
  \caption{Text examples of minority stress.}
  \begin{tabular}{p{0.9cm} p{2.6cm} p{6.9cm}} 
  \hline 
    Label & Dataset & Text \\
  \hline
    0 & Saha et al. \citep{saha2019language} & I found a cool thread [...] Thought I would post it to see other people's opinions. \\
    1 & Saha et al. \citep{saha2019language} & I hate my life, I hate being in the closet... \\
    0 & MiSSoM+ \citep{cascalheira2024lgbtq+} & I have lived with my partner for years but I have always been attracted to other women. But I'm mostly attracted to their aesthetics not sexually. \\
    1 & MiSSoM+ \citep{cascalheira2024lgbtq+} & I have to be straight if I want things in life. Being a lesbian will mean having a life where everything I want will be extremely hard to get. \\
  \hline
  \end{tabular}
\end{table}

\subsection{Interpretability and Explainability}
To better understand model behavior on psychologically sensitive text, we present three complementary interpretability analyses: (i) word cloud visualization, (ii) a word co-occurrence graph, and (iii) self-attention heatmap inspection. Together, these methods help open the “black box” of our models and validate that their decision-making aligns with known linguistic and psychological cues relevant to minority stress.

\begin{figure}[!htb]
    \centering
    \subfloat[\centering Positive]{{\includegraphics[width=5.5cm]{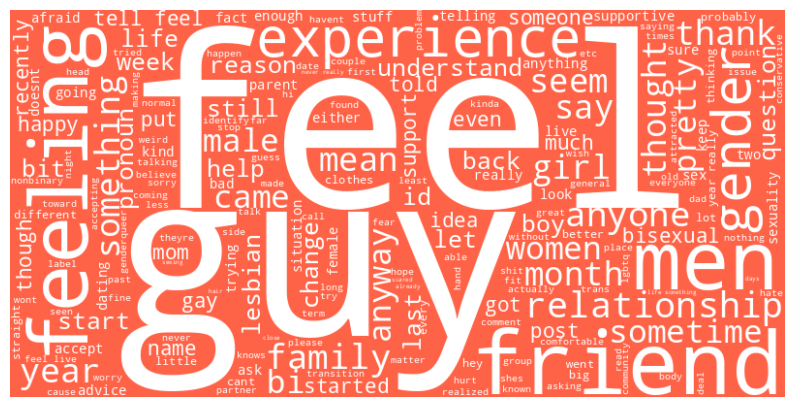} }}%
    \qquad
    \subfloat[\centering Negative]{{\includegraphics[width=5.5cm]{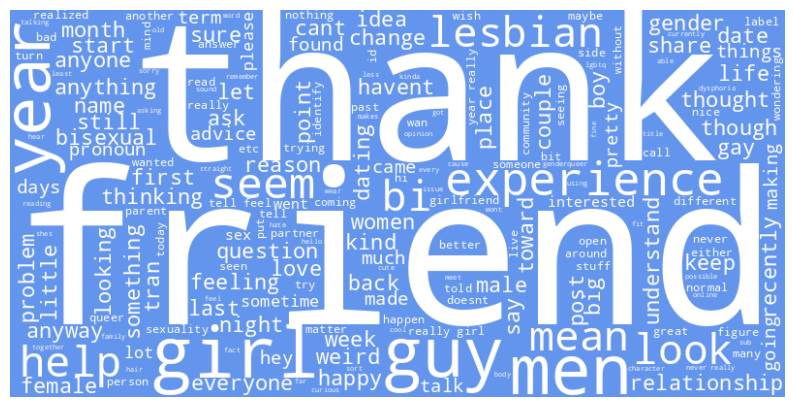} }}%
    \caption{WordCloud on LGBTQ+ MiSSoM+ dataset labels.}%
    \label{fig:example2}%
\end{figure}

Figure~\ref{fig:example2} shows word clouds for posts with and without minority stress. In positively labeled posts, salient words include \textbf{feel}, \textbf{guy}, \textbf{feeling}, \textbf{friend}, \textbf{men}, and \textbf{gender}, often occurring in syntactic frames that express personal states (e.g., first-person pronoun + affective verb). In negatively labeled posts, frequent terms include \textbf{thank}, \textbf{friend}, \textbf{girl}, \textbf{guy}, and \textbf{men}. These lexical differences not only highlight thematic divergence but also map onto semantic categories described in~\citep{cascalheira2024lgbtq+}, such as social relations, identity, and emotional expression.

\begin{figure}[htbp]  
\centerline{\includegraphics[width=0.8\linewidth]{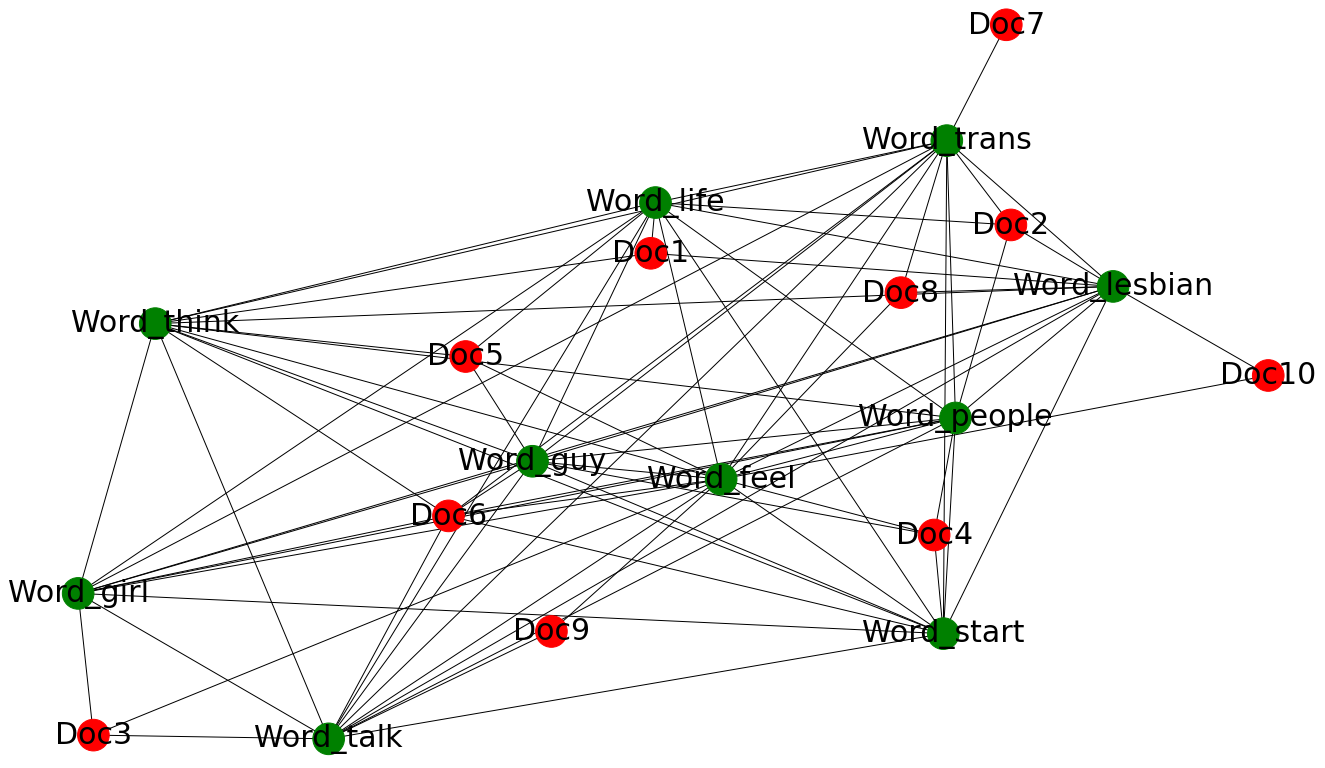}}  
\caption{Heterogeneous Graph: TF-IDF and PPMI Edge Construction from 10 Documents with Top 5 Words of MiSSoM+ dataset.}  
\label{graphArch}  
\end{figure}

The word co-occurrence graph (Figure~\ref{graphArch}) depicts ten posts (red nodes) and their most important words (green nodes). Edges connect posts to their salient words, and words to each other when they frequently co-occur. On the left, clusters around \textbf{think}, \textbf{feel}, and \textbf{guy} capture introspective and self-referential discourse. On the right, \textbf{trans}, \textbf{lesbian}, and \textbf{people} form an identity-oriented cluster. Connector words such as \textbf{talk} and \textbf{start} bridge these themes, while \textbf{life} and \textbf{girl} act as hubs linking personal feelings and social identity. These structural relationships align with the semantic networks and psychological constructs (e.g., self-identity, community affiliation) outlined in~\citep{cascalheira2024lgbtq+}.

\begin{figure}[htbp]  
\centerline{\includegraphics[width=0.7\linewidth]{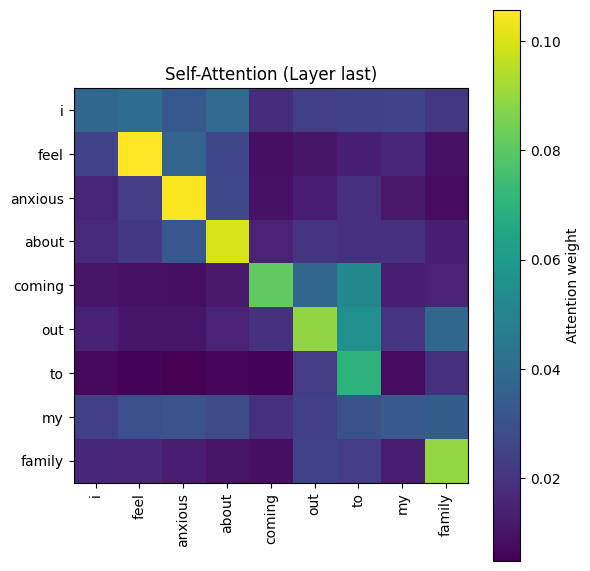}}  
\caption{Heatmap showing attention weight for positive class (minority stress).}  
\label{hmap}  
\end{figure}

The attention heatmap (Figure~\ref{hmap}) reveals that the model predominantly attends to content-bearing words rather than function words. For instance, \textbf{feel} and \textbf{anxious} are linked, signaling recognition of emotional states; \textbf{coming} and \textbf{out} are grouped and connected to \textbf{family}, capturing the culturally and psychologically loaded event of coming out to family members. Such patterns demonstrate that the model’s learned attention extends beyond isolated keywords: it captures the syntactic composition (e.g., verb–object and multiword expressions), semantic cohesion (related concepts appearing in context), and psychological salience (emotion, identity, and relational categories per~\cite{cascalheira2024lgbtq+}) of language. 

\subsection{Baseline Methods}
We compare the classification performance of our model with several other models performed by  \cite{cascalheira2023predicting} and \cite{saha2019language}:

\begin{itemize}
\item \textbf{Naive Bayes}: It is a probabilistic algorithm that applies Bayes' theorem with an assumption of independence between features \citep{b19}. The baseline was implemented using Multinomial Naive Bayes with the Term Frequency-Inverse Document Frequency (TF-IDF) \citep{c5} for feature extraction. 

\item \textbf{Logistic Regression}: It is a classification algorithm that uses a logistic function to map input variables to probabilities, determining the likelihood of belonging to a particular class \citep{b19}. It was trained using TD-IDF \citep{c5} with L2 regularization. 

\item \textbf{Support Vector Machine (SVM)}: It optimizes decision boundaries by maximizing margins between support vectors, aiming to minimize errors and enhance model generalization \citep{b19}. It was trained using TF-IDF \citep{c5} with a regularization parameter of 1 and the radial basis function kernel.

\item \textbf{Random Forest}: It is an ensemble learning method that combines decision trees by training each tree on a random subset of features. Predictions are made through a majority vote or averaging of individual tree predictions \citep{b19}. The baseline comprised 100 decision trees trained on the TF-IDF \citep{c5}.

\item \textbf{AdaBoost}: It is an ensemble learning algorithm that iteratively trains weak classifiers on different weighted versions of the training data. It assigns higher weights to misclassified samples in each iteration, allowing subsequent weak classifiers to focus on those instances \citep{b32}. The baseline was trained on base estimator of \textit{DecisionTreeClassifier} with max depth of 1, number of estimators as 50, and TF-IDF \citep{c5}.

\item \textbf{Multi-Layer Perceptron (MLP)}: It is a feed-forward neural network architecture composed of multiple layers of fully connected neurons \citep{b19}. It was trained with TF-IDF \citep{c5} using the default hyperparameters, which included a hidden layer of 100, adam as optimizer, and relu as activation function.  

\item \textbf{BiLSTM}: It is an RNN architecture that processes sequential data by incorporating both forward and backward information flow to capture contextual dependencies \citep{b33}. This model, trained with pretrained GloVe embeddings \citep{bn31}, utilized a batch size of 64, an alpha of 0.001, a hidden size of 128, a dropout rate of 0.5, and Adam optimization.

\end{itemize}

\section{Results}
\subsection{Comparative Analysis of Classification Models}
Tables~\ref{tab:saha} and~\ref{tab:lgbtq} compare the performance of classifiers on the ~\cite{saha2019language} ($N = 12{,}645$) and MiSSoM+~\citep{cascalheira2024lgbtq+} ($N = 5{,}789$) datasets using precision, recall, and F1-score. On the Saha dataset, traditional models such as SVM (F1 = 0.7401) and MLP (F1 = 0.7499) performed moderately well. Among transformer-based methods, RoBERTa(Large)-BiGRU achieved the highest F1-score (0.7718), though its improvement over the corresponding backbone (RoBERTa) was not statistically significant. This suggests that the noisy, partially machine-labeled nature of Saha may limit the benefit of architectural augmentations.  

In contrast, MiSSoM+, which is fully human-annotated by domain experts, shows consistent gains from enhanced transformer architectures. RoBERTa-GCN achieved the highest F1-score (0.8536), significantly outperforming its backbone RoBERTa ($p<0.001$). Similar patterns were observed for BERT-CNN, BERT-BiGRU, and RoBERTa-BiGRU, indicating that the cleaner, high-quality labels in MiSSoM+ enable models to exploit relational structure (via GCN/GAT) and local sequential patterns (via CNN/BiGRU) more effectively.  

\noindent\textbf{Statistical Significance Testing:}  
We assessed whether architectural enhancements significantly improve performance over their own transformer backbones using two-tailed paired $t$-tests on five independent runs with identical splits. The Bonferroni correction was applied to control for family-wise error in multiple comparisons. Significance markers in Tables~\ref{tab:saha}--\ref{tab:lgbtq} indicate improvements over the corresponding backbone: $^{\dagger}$ ($p<0.05$), $^{\ddagger}$ ($p<0.01$), $^{\star}$ ($p<0.001$).  

\begin{table}[!htb]
\centering
\caption{Comparison of classifiers' performances in ~\cite{saha2019language} ($N = 12{,}645$). Values are mean$_{\text{std}}$ over 5 runs. Significance markers denote improvements vs.\ the corresponding backbone. Bold marks highest in column.}
\begin{tabular}{p{3.5cm} p{2.5cm} p{2.5cm} p{2.5cm}}
\hline
Model & Precision & Recall & F1 \\
\hline
Naïve Bayes                 & 0.7001$_{0.012}$ & 0.4897$_{0.015}$ & 0.5302$_{0.014}$ \\
Logistic Regression         & 0.7299$_{0.010}$ & 0.7203$_{0.009}$ & 0.7199$_{0.010}$ \\
SVM                         & 0.7405$_{0.009}$ & 0.7398$_{0.022}$ & 0.7401$_{0.012}$ \\
Random Forest               & 0.7603$_{0.011}$ & 0.6701$_{0.012}$ & 0.7002$_{0.011}$ \\
AdaBoost                    & 0.7298$_{0.010}$ & 0.7204$_{0.019}$ & 0.7201$_{0.010}$ \\
MLP                         & 0.7502$_{0.011}$ & 0.7399$_{0.321}$ & 0.7499$_{0.347}$ \\
BiLSTM                      & 0.6302$_{0.332}$ & 0.6205$_{0.012}$ & 0.6101$_{0.013}$ \\
BERT-CNN                    & 0.7631$_{0.008}$ & 0.7681$_{0.451}$ & 0.7630$_{0.500}$ \\
BERT(Base)-BiGRU            & 0.7719$_{0.497}$ & 0.7626$_{0.127}$ & 0.7476$_{0.218}$ \\
BERT(Large)-BiGRU           & 0.7714$_{0.200}$ & 0.7653$_{0.186}$ & 0.7509$_{0.117}$ \\
RoBERTa(Base)-BiGRU         & 0.7829$_{0.156}$ & 0.7805$_{0.665}$ & 0.7700$_{0.425}$ \\
RoBERTa(Large)-BiGRU        & \textbf{0.7841$_{0.185}$} & \textbf{0.7813$_{0.116}$} & \textbf{0.7718$_{0.152}$} \\
\hline
\end{tabular}
\label{tab:saha}
\end{table}

\begin{table}[!htb]
\centering
\caption{Comparison of classifiers' performances in MiSSoM+~\citep{cascalheira2024lgbtq+} ($N = 5{,}789$). Values are mean$_{\text{std}}$ over 5 runs. Significance markers denote improvements vs.\ the corresponding backbone. Bold marks highest in column.}
\begin{tabular}{p{3.5cm} p{2.5cm} p{2.5cm} p{2.5cm}}
\hline
Model & Precision & Recall & F1 \\
\hline 
Naïve Bayes                  & 0.6204$_{0.058}$ & 0.7896$_{0.011}$ & 0.6905$_{0.012}$ \\
Logistic Regression          & 0.7898$_{0.222}$ & 0.8102$_{0.008}$ & 0.7801$_{0.009}$ \\
SVM                          & 0.7899$_{0.823}$ & 0.8101$_{0.118}$ & 0.7899$_{0.629}$ \\
Random Forest                & 0.7801$_{0.326}$ & 0.7899$_{0.210}$ & 0.7003$_{0.211}$ \\
AdaBoost                     & 0.7803$_{0.265}$ & 0.7897$_{0.328}$ & 0.7802$_{0.259}$ \\
MLP                          & 0.7802$_{0.211}$ & 0.7998$_{0.109}$ & 0.7804$_{0.159}$ \\
BiLSTM                       & 0.7103$_{0.186}$ & 0.7497$_{0.211}$ & 0.7197$_{0.212}$ \\
BERT-CNN                     & 0.8425$_{0.200}$ & 0.8516$_{0.006}$ & 0.8389$_{0.007}$$^{\dagger}$ \\
BERT(Base)-BiGRU             & 0.8317$_{0.033}$ & 0.8369$_{0.107}$ & 0.8302$_{0.098}$$^{\dagger}$ \\
BERT(Large)-BiGRU            & 0.8464$_{0.106}$ & 0.8530$_{0.105}$ & 0.8475$_{0.106}$$^{\ddagger}$ \\
RoBERTa(Base)-BiGRU          & 0.8306$_{0.378}$ & 0.8376$_{0.757}$ & 0.8308$_{0.568}$$^{\dagger}$ \\
RoBERTa(Large)-BiGRU         & 0.8484$_{0.654}$ & 0.8563$_{0.542}$ & 0.8478$_{0.532}$$^{\ddagger}$ \\
BERT-GCN                     & 0.8036$_{0.856}$ & 0.8198$_{0.821}$ & 0.8116$_{0.825}$ \\
RoBERTa-GCN                  & \textbf{0.8636$_{0.321}$}$^{\star}$ & \textbf{0.8586$_{0.258}$}$^{\star}$ & \textbf{0.8536$_{0.245}$}$^{\star}$ \\
BERT-GAT                     & 0.8250$_{0.125}$ & 0.8200$_{0.356}$ & 0.8150$_{0.250}$ \\
RoBERTa-GAT                  & 0.8434$_{0.236}$ & 0.8384$_{0.263}$ & 0.8334$_{0.255}$$^{\dagger}$ \\
\hline
\end{tabular}
\label{tab:lgbtq}
\end{table}

\noindent\textbf{Dataset Quality and Architectural Gains:}  
The absence of significant gains from CNN or BiGRU in the Saha dataset aligns with its partially machine-labeled nature, where annotation noise likely limits the model’s ability to exploit richer structural patterns. By contrast, MiSSoM+’s human-annotated labels provide a cleaner training signal, enabling graph-augmented (GCN/GAT) and sequentially-enhanced (CNN/BiGRU) transformer architectures to realize measurable and statistically significant improvements over their backbones.

\subsection{Performance of transformer-only models}
Table 4 shows the results on the effectiveness of transformer-based language models in detecting minority stress in the discourse on social networks. The dataset was split using a 70:15:15 ratio, where 70\% was allocated for training, and the remaining 30\% was further divided into validation sets (15\%) and test sets (15\%) using stratified sampling to preserve the distribution of the label. We utilized the base model for all transformers to maintain a fair comparison. Among the models tested on the MiSSoM+ dataset, BART achieved the highest performance (F1 score: 0.8586, accuracy: 0.8550), with RoBERTa and Longformer also performing well. GPT-2 had the lowest F1 score (0.6918) but maintained a high recall (0.7860). On the Saha et al. dataset, BERT emerged as the best performer (F1 score: 0.8397, accuracy: 0.8412), followed by DeBERTa and ELECTRA, while GPT-2 remained the weakest. Similarly, Table 5 shows the results of the zero-shot and few-shot learning experiments, where ChatGPT (10-shot) emerged as the top performer (F1 score: 0.7437, accuracy: 0.7963), performing better than zero-shot settings and 3-shot settings, as well as most fine-tuned traditional machine learning models.

\begin{table}[!htb]
    \centering
    \caption{Performance evaluation of transformer-only models on the MiSSoM+ \citep{cascalheira2024lgbtq+} and \cite{saha2019language} datasets. Bold marks highest in column for each dataset.}
    \begin{tabular}{p{1.8cm} p{1.8cm} p{1.8cm} p{1.8cm} p{1.8cm} p{1.2cm}}
        \hline
        Dataset     & Model       & Precision        & Recall           & F1               & Accuracy         \\
        \hline
        \multirow{7}{*}{MiSSoM+} 
        & DeBERTa     & 0.8403$_{0.189}$ & 0.8423$_{0.172}$ & 0.8412$_{0.181}$ & 0.8423$_{0.172}$ \\
        & BERT        & 0.8102$_{0.221}$ & 0.8112$_{0.205}$ & 0.8106$_{0.213}$ & 0.8112$_{0.205}$ \\
        & RoBERTa     & 0.8578$_{0.147}$ & 0.8539$_{0.156}$ & 0.8556$_{0.151}$ & 0.8539$_{0.156}$ \\
        & GPT-2       & 0.6177$_{0.302}$ & 0.7860$_{0.290}$ & 0.6918$_{0.296}$ & 0.7860$_{0.290}$ \\
        & Longformer  & 0.8530$_{0.134}$ & 0.8516$_{0.128}$ & 0.8523$_{0.131}$ & 0.8516$_{0.128}$ \\
        & ELECTRA     & 0.8504$_{0.178}$ & \textbf{0.8670$_{0.162}$} & 0.8485$_{0.169}$ & \textbf{0.8670$_{0.162}$} \\
        & BART        & \textbf{0.8642$_{0.154}$} & 0.8550$_{0.140}$ & \textbf{0.8586$_{0.147}$} & 0.8550$_{0.140}$ \\
        \hline
        \multirow{7}{*}{Saha et al.} 
        & DeBERTa     & 0.8348$_{0.112}$ & 0.8338$_{0.105}$ & 0.8342$_{0.108}$ & 0.8338$_{0.105}$ \\
        & BERT        & \textbf{0.8393$_{0.098}$} & \textbf{0.8412$_{0.092}$} & \textbf{0.8397$_{0.095}$} & \textbf{0.8412$_{0.092}$} \\
        & RoBERTa     & 0.7977$_{0.125}$ & 0.7983$_{0.119}$ & 0.7980$_{0.122}$ & 0.7983$_{0.119}$ \\
        & GPT-2       & 0.4309$_{0.276}$ & 0.6564$_{0.263}$ & 0.5203$_{0.269}$ & 0.6564$_{0.263}$ \\
        & Longformer  & 0.7607$_{0.147}$ & 0.7665$_{0.139}$ & 0.7611$_{0.143}$ & 0.7665$_{0.139}$ \\
        & ELECTRA     & 0.8331$_{0.104}$ & 0.8354$_{0.099}$ & 0.8331$_{0.101}$ & 0.8354$_{0.099}$ \\
        & BART        & 0.8200$_{0.118}$ & 0.8221$_{0.111}$ & 0.8206$_{0.114}$ & 0.8221$_{0.111}$ \\
        \hline
    \end{tabular}
    \label{tab:combined_transformer_performance}
\end{table}

\begin{figure}[!htb]
    \centering
    \subfloat[\centering]{{\includegraphics[width=6cm]{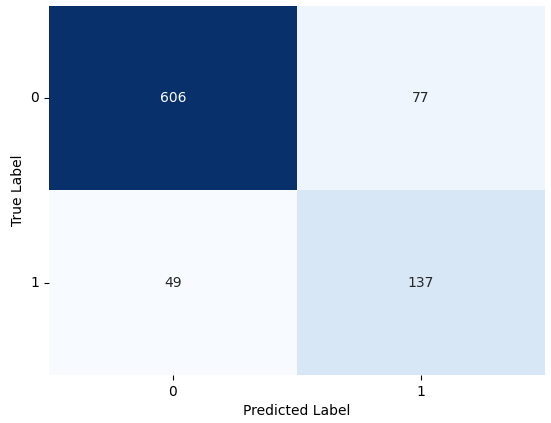} }}%
    \qquad
    \subfloat[\centering]{{\includegraphics[width=6cm]{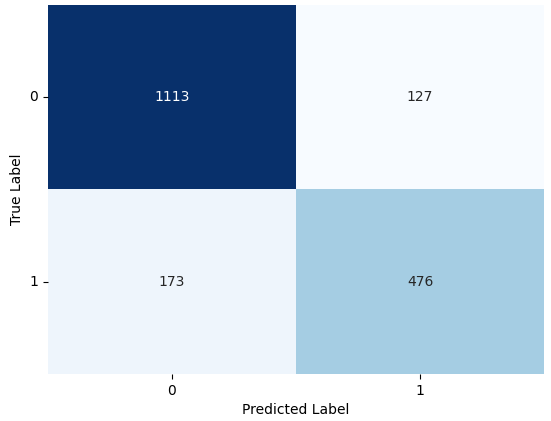} }}%
    \caption{Confusion Matrix of BART on (a) MiSSoM+ \citep{cascalheira2024lgbtq+} and BERT on (b) \cite{saha2019language}}%
    \label{fig:conf}%
\end{figure}

Figure 8 shows the confusion matrices of BART in the MiSSoM+ dataset \citep{cascalheira2024lgbtq+} and BERT in the \cite{saha2019language} dataset showing their classification performance on two different datasets. A direct comparison between BART and BERT is challenging due to differences in their datasets. However, analyzing each model on its respective dataset can be valuable for assessing performance. 

BART performs well in identifying negative cases, but struggles in detecting actual minority stress, leading to a high false negative rate and lower recall in MiSSoM+ dataset \citep{cascalheira2024lgbtq+}. On the other hand, BERT shows a better balance between precision and recall, capturing more minority stress cases with a lower false negative rate in the \cite{saha2019language} dataset. These results suggest that BERT may generalize better for stress detection in its dataset, whereas BART might require improvements to improve minority stress detection. 

\begin{figure*}[htbp]
\centerline{\includegraphics[width=0.5\textwidth]{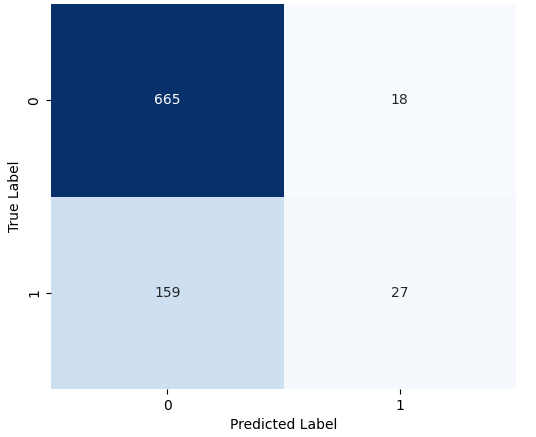}}
\caption{Confusion Matrix of ChatGPT 10-shot on MiSSoM+ \citep{cascalheira2024lgbtq+}}
\label{chatg}
\end{figure*}

\subsection{Few-Shot Performance}

\begin{table}[!htb]
    \centering
    \caption{Zero-Shot and Few-Shot results in 869 samples of MiSSoM+ \citep{cascalheira2024lgbtq+} (results as of January 26, 2025). Bold marks highest in column.}
    \begin{tabular}{p{3cm} p{2cm} p{2cm} p{2cm} p{2cm}}
        \hline
        Model                  & Precision       & Recall          & F1              & Accuracy        \\
        \hline
        RoBERTa (Zero-shot)    & 0.6427$_{1.234}$ & 0.6110$_{2.001}$ & 0.6257$_{1.578}$ & 0.6110$_{2.001}$ \\
        ChatGPT (Zero-shot)    & 0.7015$_{1.847}$ & 0.6122$_{2.315}$ & 0.6432$_{1.429}$ & 0.6122$_{2.315}$ \\
        ChatGPT (3-shot)       & 0.7345$_{1.102}$ & 0.7445$_{1.907}$ & 0.7391$_{1.256}$ & 0.7445$_{1.907}$ \\
        ChatGPT (10-shot)      & \textbf{0.7627}$_{2.781}$ & \textbf{0.7963}$_{1.654}$ & \textbf{0.7437}$_{2.023}$ & \textbf{0.7963}$_{1.654}$ \\
        \hline
    \end{tabular}
    \label{tab:zero_few_shot}
\end{table}

The ChatGPT 10-shot classification model performed well in identifying non-minority stress cases, but struggled with detecting minority stress. The confusion matrix (Figure 9) shows 665 True Negatives and 27 True Positives, but a high False Negative count (159) indicates that many minority stress cases were missed. Although False positives were low (18), the model's recall is weaker than precision, meaning it correctly identifies minority stress when predicted but often fails to detect it. This explains why the weighted F1 score (0.7437) is lower than precision and recall. For the 10-shot classification, we performed an error analysis on the misclassified samples and found that most errors were due to indirect context, idiomatic language, sarcasm, and similar phenomena. Table 6 summarizes these error categories with descriptions and illustrative examples of ChatGPT’s performance. These results demonstrate the potential of AI-driven health interventions to create personalized stress reduction platforms, similarly reinforcing AI’s role in digital health and public policy.

\begin{table}[htbp]
\centering
\caption{Qualitative error categories for misclassified minority stress expressions in ChatGPT’s 10-shot classification}

\begin{tabular}{p{2.6cm} p{4.2cm} p{4cm}}
\hline
\textbf{Category} & \textbf{Description} & \textbf{Example} \\
\hline
Indirect Context & Stress expressed via situation references without explicit keywords; depends on prior posts &  
“I finally told them last week… now I can’t sleep.” \\
\addlinespace
Idiomatic Language & Community-specific idioms not seen in training; literal reading fails &  
“I’m keeping it in the vault for now.” (i.e.\ “in the closet”) \\
\addlinespace
Sarcasm / Irony & Polarity flipped by tone; stress masked under humor or irony &  
“Sure, because everyone loves me being queer… *eyeroll*” \\
\addlinespace
Mixed Sentiment & Posts combine positive and negative affect, confusing binary labeler &  
“I feel great today, but I still dread tomorrow’s outing.” \\
\addlinespace
Implicit Stigma & Self-directed negative statements without explicit “hate” keywords &  
“Sometimes I wish I’d never realized this about myself.” \\
\hline
\end{tabular}
\label{tab:qual_errors}
\end{table}

\subsection{Ablation Study}

\(\lambda\) is the hyperparameter in BERT–GCN that balances the contributions of the BERT and GCN modules. Varying \(\lambda\) produces different accuracy and F1‐score results. Figure 10 shows classification performance on the LGBTQ\(+\) MiSSoM\(+\) test set across several \(\lambda\) values. Setting \(\lambda = 0\) uses only the BERT module, while \(\lambda = 1\) uses only the GCN. Through extensive experiments, we found that \(\lambda = 0.2\) yields the best trade‐off, achieving peak accuracy and F1‐score. For RoBERTa, integrating the GCN boosts accuracy by \(1.97\%\) and F1‐score by \(1.44\%\). Although BERT pretraining generally contributes more than graph learning, appropriately weighting and incorporating the GCN component can still deliver a meaningful accuracy improvement. Figure 11 shows the percentage changes observed in our ablation study. Jointly deploying BERT and the CNN clearly outperforms either model alone. Recall sees the largest gain—up \(5.43\%\) compared to BERT only—while overall accuracy improves by \(4.57\%\) and F1‐score by \(2.43\%\). Precision remains largely unchanged. These results confirm that combining BERT with a convolutional graph encoder enhances the model’s overall classification performance.

\begin{figure}[!htb]
    \centering
    \subfloat[\centering Accuracy and F1 score with varying $\lambda$ on RoBERTa-GCN.]{{\includegraphics[width=7cm]{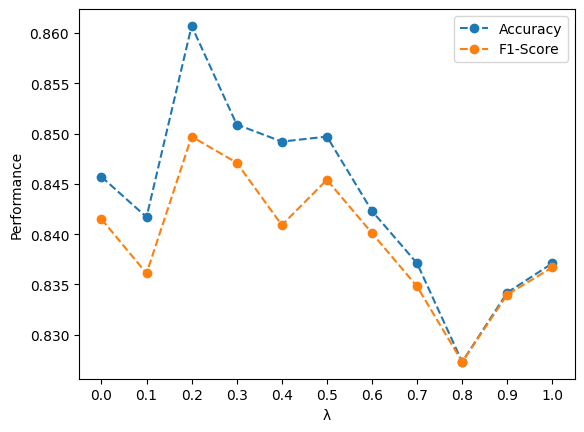}}}%
    \qquad
    \subfloat[\centering Performance difference from ablation study on BERT-CNN.]{{\includegraphics[width=7cm]{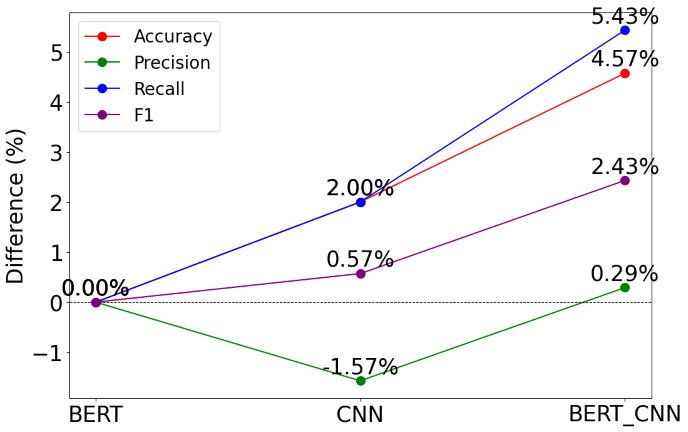}}}%
    \caption{Ablation analyses on the LGBTQ+ MiSSoM+ dataset.}%
    \label{fig:ablation_combined}%
\end{figure}

\section{DISCUSSION}
Our results strongly support the central hypothesis that augmenting transformer models with relational or structural context improves minority stress detection. Experiments focused primarily on the MiSSoM+ dataset, which contains high-quality human-annotated labels, ensuring reliable evaluation. On this dataset, graph-augmented models such as RoBERTa-GCN and BERT-GCN outperformed their sequence-only counterparts, confirming that social and semantic structure enhances model sensitivity to certain stress expressions like identity concealment and internalized stigma. We also found that integrating convolutional layers into BERT (i.e., BERT-CNN) improved performance by capturing local text patterns that are present in stress-related language. These enhancements demonstrate that both graph and CNN-based architectural additions contribute meaningful gains over standard transformers. While models like RoBERTa and BART performed well on their own, graph-augmented and CNN-enhanced versions consistently delivered superior results. In contrast, zero- and few-shot models like ChatGPT showed promise but lagged behind supervised approaches, particularly in recall.

In summary, our findings validate that combining transformers with relational and local modeling leads to more accurate detection of minority stress. These results, grounded in the MiSSoM+ dataset, highlight the importance of domain-specific tuning and structured context for effective digital health interventions.

\subsection{Digital Health Interventions}
As noted in past research \citep{cascalheira2024lgbtq+}, the accurate prediction of minority stress could inform digital health interventions. Interventionists could develop a personalized stress reduction smartphone app for minority people that triggers a notification reminder to use a coping skill when minority stress is detected. Furthermore, in preexisting digital health interventions, app developers can use our models to adjust content shown in a modular mobile health app given the presence or absence of minority stress. For example, if a minority person is starting a course of computer-delivered therapy for stress-induced health conditions, delivering modules on distress tolerance during high-stress periods may be superior to delivering modules in a linear fashion. Highly accurate detection of minority stress could also influence the timing of booster sessions (i.e., follow-up therapy sessions provided after a course of psychotherapy). If an app built with our models detects minority stress at a given threshold, a reminder could be sent to the client or therapist to prompt outreach. Each of these downstream applications has the potential to reduce minority stress for minority individuals. (e.g., ,expressive writing and self-affirmation interventions).

\subsection{Public Health and Policy}
Beyond enhancing personal health, our results pave the way toward more effective public health and policy. Our findings could be used to forecast possible surges in minority stress following specific events (e.g., outbreaks of new sexually transmitted diseases, gun violence in nightclubs); such predictions could lead to better allocation of resources across minority community spaces. In addition, policymakers and social scientists could track how the passage of specific legislation influences levels of minority stress , which can be difficult to quantify and demonstrate as effective. For example, state policies that effectively reduce minority stress could be replicable in new jurisdictions, while policies with fewer reductions in minority stress might be revised. This measurement-based approach to policy making could assist advocates and other stakeholders in prioritizing actions with the largest reduction in minority stress.

\subsection{Limitations and Future Research}
Although our experiments resulted in excellent performance, there are several possible next steps to improve the classification of minority stress. The data used were sourced from Reddit, which may not fully capture the experiences of the broader minority community. To ensure that diverse linguistic patterns and stress expressions are considered, future research should explore other social media platforms like Twitter and TikTok. From a social science perspective, we utilized a human-annotated dataset grounded in minority stress theory. Although this theory is widely recognized for explaining the social stressors faced by minority individuals, we acknowledge its limitations in addressing the complexities of intersecting identities (e.g., being both a gender/sexual minority and belonging to a low-income socioeconomic group) \citep{rivas2023temporal, noyola2020minority}. Future research should further explore minority stress by incorporating a more comprehensive understanding of social identities and their intersections. Finally, although model performance metrics are promising, their implementation in healthcare or policy settings requires further validation. Future studies should assess how well these models function in real-world applications, including integration into digital mental health services. Future work should explore additional robustness strategies. Pruning-based defenses in large language models show promise for mitigating hidden vulnerabilities \citep{chapagain2025pruning}. Integrating pruning with augmentation could enhance both accuracy and resilience in minority stress detection.

\subsection{Ethical Consideration}
While our study adheres to standard practices for analyzing public Reddit data, we recognize deeper ethical complexities in working with stigmatized communities and sensitive topics like minority stress. Users may not anticipate that their disclosures, shared in vulnerable moments, could be subject to computational analysis. Although we removed all identifying information and restricted data access, the lack of informed consent remains ethically significant. For this reason, we restrict access to the datasets in this paper to protect user privacy, releasing only the code and Python scripts necessary to reproduce our analyses. Interested and qualified scientists should contact us directly for data access. Our models also carry risks in potential real-world applications. False negatives could result in missed opportunities for support, while false positives may misrepresent needs or reinforce stereotypes. These misclassifications can have real consequences, especially for marginalized populations. We caution against deploying such models in clinical or policy settings without rigorous validation and community involvement. Finally, we acknowledge that our work reflects the assumptions of minority stress theory, which may not capture the full complexity of intersecting identities. Future research should pursue more participatory, intersectional approaches and deepen ethical engagement beyond procedural norms.

\section{Acknowledgements}
Cory J. Cascalheira is supported as a RISE Fellow by the National Institutes of Health (R25GM061222). Shah Muhammad Hamdi is supported by the CISE and GEO directorates under NSF awards \#2301397 and \#2305781. Soukaina Filali Boubrahimi is supported by CISE and GEO Directorates under NSF awards \#2204363, \#2240022, \#2301397, and \#2305781. Jillian R. Scheer is supported by a Mentored Scientist Development Award (K01AA028239-01A1) from the National Institute on Alcohol Abuse and Alcoholism.

\bibliography{sn-bibliography}

\begin{appendices}

\section{Checklist}\label{secA1}

\begin{enumerate}
    \item For most authors...
    
    \begin{enumerate}
        \item Would answering this research question advance science without violating social contracts, such as violating privacy norms, perpetuating unfair profiling, exacerbating the socio-economic divide, or implying disrespect to societies or cultures? \textcolor{blue}{Yes, see Ethical Considerations section.}
        \item Do your main claims in the abstract and introduction accurately reflect the paper's contributions and scope? \textcolor{blue}{Yes, see Abstract and Introduction.}
        \item Do you clarify how the proposed methodological approach is appropriate for the claims made? \textcolor{blue}{Yes, see Introduction and Related Work.}
        \item Do you clarify what are possible artifacts in the data used, given population-specific distributions? \textcolor{blue}{Yes, see Limitations and Future Research.}
        \item Did you describe the limitations of your work? \textcolor{blue}{Yes, see Limitations and Future Research.}
        \item Did you discuss any potential negative societal impacts of your work? \textcolor{blue}{Yes, see Ethical Considerations section.}
        \item Did you discuss any potential misuse of your work? \textcolor{blue}{Yes, see Ethical Considerations section.}
        \item Did you describe steps taken to prevent or mitigate potential negative outcomes of the research, such as data and model documentation, data anonymization, responsible release, access control, and the reproducibility of findings? \textcolor{blue}{Yes, see Ethical Considerations section.}
        \item Have you read the ethics review guidelines and ensured that your paper conforms to them? \textcolor{blue}{Yes.}
    \end{enumerate}

    \item Additionally, if your study involves hypotheses testing...

    \begin{enumerate}
        \item Did you clearly state the assumptions underlying all theoretical results? \textcolor{blue}{N/A}
        \item Have you provided justifications for all theoretical results? \textcolor{blue}{N/A}
        \item Did you discuss competing hypotheses or theories that might challenge or complement your theoretical results? \textcolor{blue}{N/A}
        \item Have you considered alternative mechanisms or explanations that might account for the same outcomes observed in your study? \textcolor{blue}{N/A}
        \item Did you address potential biases or limitations in your theoretical framework? \textcolor{blue}{No, critiquing the minority stress framework is too large of a task for a paper of the current scope (i.e., where the purpose was to improve prediction of minority stress on social media).}
        \item Have you related your theoretical results to the existing literature in social science? \textcolor{blue}{N/A}
        \item Did you discuss the implications of your theoretical results for policy, practice, or further research in the social science domain? \textcolor{blue}{N/A}
    \end{enumerate}

    \item Additionally, if you are including theoretical proofs...
     \begin{enumerate}
        \item Did you state the full set of assumptions of all theoretical results?  \textcolor{blue}{N/A}
        \item Did you include complete proofs of all theoretical results? \textcolor{blue}{N/A}
    \end{enumerate}

    \item Additionally, if you ran machine learning experiments...
    \begin{enumerate}
        \item Did you include the code, data, and instructions needed to reproduce the main experimental results (either in the supplemental material or as a URL)? \textcolor{blue}{Yes, we include a link to all Python scripts. However, we do not release the data—see Ethical Considerations for more details.}
        \item Did you specify all the training details (e.g., data splits, hyperparameters, how they were chosen)? \textcolor{blue}{Yes, see Methodology and Experiments.}

        \item Did you report error bars (e.g., with respect to the random seed after running experiments multiple times)? \textcolor{blue}{No, however, we reported the mean metric after five independent trials with unique random seeds.} 
        \item Did you include the total amount of compute and the type of resources used (e.g., type of GPUs, internal cluster, or cloud provider)? \textcolor{blue}{Yes, see Experiments.}

        \item Do you justify how the proposed evaluation is sufficient and appropriate to the claims made? \textcolor{blue} {Yes, see Experiments.}

        \item Do you discuss what is “the cost“ of misclassification and fault (in)tolerance? \textcolor{blue}{Yes, see Results.}

    \end{enumerate}

    \item Additionally, if you are using existing assets (e.g., code, data, models) or curating/releasing new assets...
    \begin{enumerate}
        \item If your work uses existing assets, did you cite the creators? \textcolor{blue}{Yes, see Datasets.}
        \item Did you mention the license of the assets? \textcolor{blue}{Yes, see Datasets.}
        \item Did you include any new assets in the supplemental material or as a URL? \textcolor{blue}{N/A}
        \item Did you discuss whether and how consent was obtained from people whose data you're using/curating? \textcolor{blue}{Yes, see Datasets and Ethical Considerations.}
        \item Did you discuss whether the data you are using/curating contains personally identifiable information or offensive content? \textcolor{blue}{Yes, see Datasets.}
        \item If you are curating or releasing new datasets, did you discuss how you intend to make your datasets FAIR (see FORCE11 (2020))? \textcolor{blue}{N/A}
        \item If you are curating or releasing new datasets, did you create a Datasheet for the Dataset (see Gebru et al. (2021))? \textcolor{blue}{N/A}
    \end{enumerate}

    \item Additionally, if you used crowdsourcing or conducted research with human subjects...

    \begin{enumerate}
        \item Did you include the full text of instructions given to participants and screenshots? \textcolor{blue}{N/A}
        \item Did you describe any potential participant risks, with mentions of Institutional Review Board (IRB) approvals? \textcolor{blue}{Yes, see Experiments.}
        \item Did you include the estimated hourly wage paid to participants and the total amount spent on participant compensation? \textcolor{blue}{N/A}
        \item Did you discuss how data is stored, shared, and deidentified? \textcolor{blue}{Yes, see Datasets.}

    \end{enumerate}
\end{enumerate}

\end{appendices}

\end{document}